# OPTIMIZER SENSITIVITY IN VISION TRANSFORMER-BASED IRIS RECOGNITION: ADAMW VS SGD VS RMSPROP

Moh. Imam Faiz[1], Aviv Yuniar Rahman[2], Rangga Pahlevi Putra

Informatics Engineering, Universitas Widya Gama Malang

[1]faizmeza03@gmail.com, [2]Aviv@widyagama.ac.id, [3]rangga@widyagama.ac.id

**Abstract**

The security and reliability of biometric authentication have become increasingly crucial (Jain, Nandakumar, and Ross 2016) with the growing demand for digital identity systems. Among various biometric modalities, iris recognition stands out due to its unique and stable texture patterns over a person's lifetime. Recent advances in deep learning, particularly Vision Transformers (ViT), have demonstrated strong performance in visual recognition tasks. However, the sensitivity of ViT to optimizer selection in biometric applications remains underexplored.

This study systematically compares three popular optimizer AdamW, Stochastic Gradient Descent (SGD), and RMSprop—for training a ViT-Small model on the MMU-Iris dataset under a closed-set identification protocol. All models share the same architecture, data preprocessing pipeline, and training hyperparameters, allowing us to isolate the impact of the optimizer on convergence behavior and recognition performance.

The results reveal a clear performance gap between optimizers. AdamW achieves 97.78% validation accuracy and 98.89% test accuracy, with a nearly perfect confusion matrix and stable convergence. In contrast, SGD reaches only 13.33% test accuracy due to underfitting, while RMSprop fails to converge under mixed precision, exhibiting numerical instability and near-random predictions. These findings demonstrate that optimizer choice is a critical factor for ViT-based iris recognition, especially on small biometric datasets. To the best of our knowledge, this work is among the first to provide a systematic analysis of optimizer sensitivity in Vision Transformer–based iris recognition and establishes AdamW as a strong baseline for real-world biometric authentication systems.

# 1 Introduction

In the era of rapid digital transformation, the demand for secure and accurate biometric authentication has become increasingly critical across various sectors, including finance, healthcare, and border security. Biometric technologies offer a robust solution for reliable identity verification by leveraging unique physiological or behavioral traits (Jain, Nandakumar, and Ross 2016). Among the various biometric modalities, iris recognition stands out as one of the most secure and reliable approaches due to the uniqueness, complexity, and long-term stability of the iris pattern throughout a person's lifetime (Daugman 2009).

Over the years, iris recognition methods have undergone a significant evolution. Early systems relied on handcrafted features and traditional image processing methods such as Gabor and Log-Gabor filters With the rise of deep learning, Convolutional Neural Networks (CNNs) improved feature extraction and classification performance in iris recognition task ( Rizgar et al., 2025). More recently, Transformer-based architectures, particularly Vision Transformers (ViT), have demonstrated strong capabilities in capturing global contextual information in image data, outperforming CNNs in various computer vision benchmarks (Dosovitskiy et al. 2021; Liu et al. 2021)

Despite these advancements, a crucial research question remains insufficiently explored: how different optimization algorithms affect the convergence, training stability, and recognition accuracy of ViT models in biometric applications. Previous iris recognition studies have predominantly focused on architectural improvements or feature engineering, while the impact of optimizer selection has received little systematic investigation (Zhang et al. 2022).This gap is particularly relevant for small-scale biometric datasets such as MMU-Iris, where training stability can significantly influence model performance (Kadhim et al. 2025).

This research addresses that gap by systematically evaluating the performance of three popular optimizers—AdamW, SGD, and RMSprop—in training Vision Transformer models for iris recognition. Using the MMU-Iris dataset, we analyze their effects on convergence, training stability, and classification accuracy. The experimental results demonstrate a clear performance difference among optimizers, with AdamW achieving 97.78% validation accuracy and 98.89% test accuracy, while SGD and RMSprop exhibited poor convergence and unstable training dynamics.

These findings highlight the importance of optimizer selection when deploying ViT models in real-world biometric systems, particularly under data-scarce scenarios.

## 2 Related Work

### 2.1 Iris Recognition

Iris recognition has long been considered one of the most reliable and secure biometric modalities because of its unique and stable texture patterns over a person's lifetime. Traditional iris recognition systems were pioneered by Daugman, who introduced Gabor and Log-Gabor filters for efficient feature extraction and matching, enabling accurate identity verification even under moderate variations in acquisition conditions (Daugman 2009).

With the advent of deep learning, CNN-based architectures such as VGG and ResNet have surpassed handcrafted methods by learning hierarchical feature representations directly from raw images. These models demonstrate greater robustness against variations in illumination, pupil dilation, and rotation, which previously posed significant challenges to classical algorithms (Nguyen, et al. 2024)This evolution has established CNNs as the de facto baseline in modern iris recognition research.

However, despite the impressive progress, CNNs still face limitations in modeling long-range spatial dependencies(Nguyen, et al. 2024) and global texture patterns inherent in iris structures. This limitation opens opportunities for more advanced architectures capable of capturing global contextual information, such as Vision Transformers.

Research gap: While CNN-based iris recognition has been extensively explored, studies focusing on Transformer-based architectures remain limited, especially in scenarios involving small-scale datasets.

### 2.2 Vision Transformer in Biometrics

The introduction of Vision Transformers (ViT) has reshaped the landscape of computer vision by replacing local convolution operations with self-attention mechanisms, enabling the model to capture long-range dependencies across image patches (Dosovitskiy et al. 2021). Unlike CNNs, ViTs process images as sequences of patches, making them more flexible and capable of learning global patterns (Touvron et al. 2021).

In biometric recognition, Transformer architectures have shown promising results in tasks such as face recognition and palmprint verification (Sharma et al.

2025). Beyond these modalities, Vision Transformers have also been explored in fingerprint recognition. (Qiu et al. 2025) introduced IFViT, a ViT-based fingerprint matching framework, demonstrating that Transformer models are capable of learning stable and discriminative features across diverse biometric modalities. Recent studies have further applied ViT directly to iris recognition in practical scenarios. For example, (Ennajar and Bouarifi 2024) implemented a ViT-L16 model for iris-based student attendance monitoring and achieved an accuracy of 95.69%, indicating that ViT is effective not only for controlled laboratory settings but also for real-world biometric applications. These developments collectively highlight the strong potential of Vision Transformers for various biometric tasks, including iris recognition.

However, most of the existing studies focus on large-scale datasets or other biometric modalities. The behavior of ViTs on small, domain-specific biometric datasets like MMU-Iris has not been thoroughly investigated, particularly regarding training stability and generalization.

Research gap: There is a clear lack of systematic evaluation of ViT performance for iris recognition, especially under limited data conditions, which reflects many real-world biometric deployment scenarios.

### 2.3 Optimizer Impact Studies

The choice of optimizer plays a crucial role in determining the convergence behavior, training stability, and final performance of deep learning models. Adaptive optimizers such as Adam and its variant AdamW have been shown to converge faster and more stably compared to classical SGD or RMSprop, particularly in Transformer architectures (Kingma and Ba 2015). AdamW's decoupled weight decay further improves generalization by preventing overfitting.

While these findings are well documented in natural language processing and ImageNet-scale computer vision tasks, their applicability to biometric domains has not been sufficiently tested. Previous works on iris recognition have mainly focused on model architectures rather than the impact of training optimization strategies (Zhang et al. 2022)

Research gap: There is little empirical evidence comparing optimizer sensitivity for ViT in iris biometrics. This study directly addresses this gap by systematically evaluating AdamW, SGD, and RMSprop on the MMU-Iris dataset and demonstrating their effects on convergence and recognition accuracy.

## 3 Methodology

### 3.1 *Dataset and Preprocessing*

Figure 1 Vision Transformer training and evaluation process flow for iris recognition.

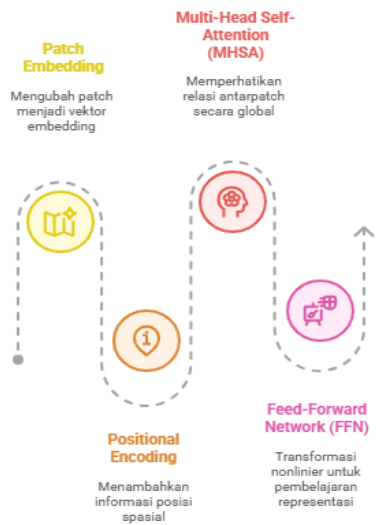

The research process flow used in this study is shown in Figure 1. This study employs the MMU Iris Database, a well-known benchmark dataset for iris recognition research. The dataset contains images from 45 subjects, with a total of 450 samples (left and right eyes) captured under controlled indoor conditions. These images include natural variations in illumination, pupil dilation, and slight head movements.

For experimental evaluation, a closed-set protocol was used with a per-subject and per-eye split: 3 training samples, 1 validation sample, and 1 test sample per class. This corresponds to a 60/20/20 train-validation-test split.

All images were converted to RGB 3-channel format and resized to 224 × 224 pixels to match the input size requirement of Vision Transformer models (Dosovitskiy et al. 2021). Normalization used the ImageNet mean and standard deviation. Data augmentation was deliberately lightweight to avoid altering iris patterns, consisting of ColorJitter (p = 0.3) and GaussianBlur (p = 0.3).

### 3.2 **Model Architecture**

Figure 2 The Vision Transformer (ViT-Small) architecture used in this study

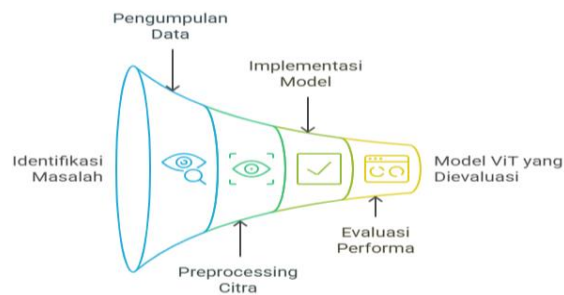

The model architecture used in this study is shown in patches (Dosovitskiy et al. 2021), enabling it to model long-range spatial dependencies via multi-head self-attention.

Figure *2*. The experiments utilized the ViT-Small architecture implemented through the Timm library. ViT processes images as sequences of 16 × 16 pixel

The configuration used in this study includes:

- 12 Transformer encoder layers,
- 6 attention heads per layer,
- Hidden dimension of 384,
- Classification performed through the [CLS] token.

To enhance convergence, pretrained weights on ImageNet were used, followed by fine-tuning on the MMU-Iris dataset. This transfer learning approach is particularly beneficial in low-data scenarios.

### 3.3 Training Strategies

The training pipeline consisted of two phases:

1. Head freezing: Training only the classification head to stabilize initial learning.
2. Full fine-tuning: Unfreezing all layers for joint optimization.

All models were trained using mixed precision (AMP) to reduce memory usage and training time. Cosine learning rate scheduling was applied, and early stopping was used to avoid overfitting (Touvron et al. 2021). The batch size was set to 16, and all other training parameters were kept constant across experiments to ensure fair comparison.

### 3.4 Optimizer Configuration

Three optimizers were compared: AdamW, SGD, and RMSprop.

Each was carefully configured to isolate the effect of the optimizer itself on model convergence and final performance

Table 1 Optimizer configuration parameters used for ViT-Small training.

| Optimizer | LR (Head) | LR (Full) | Scheduler | Momentum | Weight Decay | Notes |
|---|---|---|---|---|---|---|
| AdamW | 1e-4 | 5e-4 | Cosine | - | 0.05 | Stable convergence |
| SGD(Nesterov) | 1e-3 | 2e-4 | Cosine | 0.9 | 5e-4 | Underfitting observed |
| RMSprop(centered) | 1e-3 | 2e-4 | Cosine | 0.9 | 0 | Numerical instability (NaN) |

Table *1* was applied during training. Warm-up was not applied in the baseline setting but is recommended for SGD and RMSprop in future work.

## 4 Experiments & Evaluation

### 4.1 Evaluation Metrics

These evaluation criteria follow widely used deep learning performance standards

- Accuracy — measuring overall classification performance.
- F1-score — providing a balanced view of precision and recall, particularly useful for small or uneven datasets.
- Confusion Matrix — visualizing class-level performance and identifying misclassification patterns across the 45 iris classes.

These metrics were selected to comprehensively assess how optimizer choice impacts both model stability and recognition accuracy

Table *2*. AdamW showed clear superiority, achieving the highest validation and test accuracy with stable

### 4.2 Experimental Setup

All experiments were performed on a workstation equipped with an NVIDIA RTX GPU (32 GB), running PyTorch with deterministic seeds to ensure reproducibility.

- Mixed Precision (AMP) was used to accelerate training while maintaining numerical stability.
- Early stopping was applied to prevent overfitting.
- Training was conducted for up to 100 epochs with Cosine learning rate scheduling.
- All optimizer configurations shared the same batch size (16) and label smoothing (0.1) to isolate the effect of the optimizer itself.

### 4.3 Results Comparison

The results of the optimizer comparison are summarized in convergence, while SGD suffered from underfitting and RMSprop experienced training instability (NaN losses).

Table 2 Optimizer performance on MMU-Iris (Closed-Set)

| Optimizer | Val Acc | Test Acc | F1-score | Observation |
| --- | --- | --- | --- | --- |

| AdamW | 0.9778 | 0.9889 | 0.98 | Stable training |
| SGD (Nesterov) | 0.1222 | 0.1333 | 0.13 | Underfitting |
| RMSprop (centered) | 0.0222 | 0.0222 | 0.02 | NaN instability |

## 4.4 confusion Matrix Analysis

Figure 3 Confusion matrix of test results using the AdamW

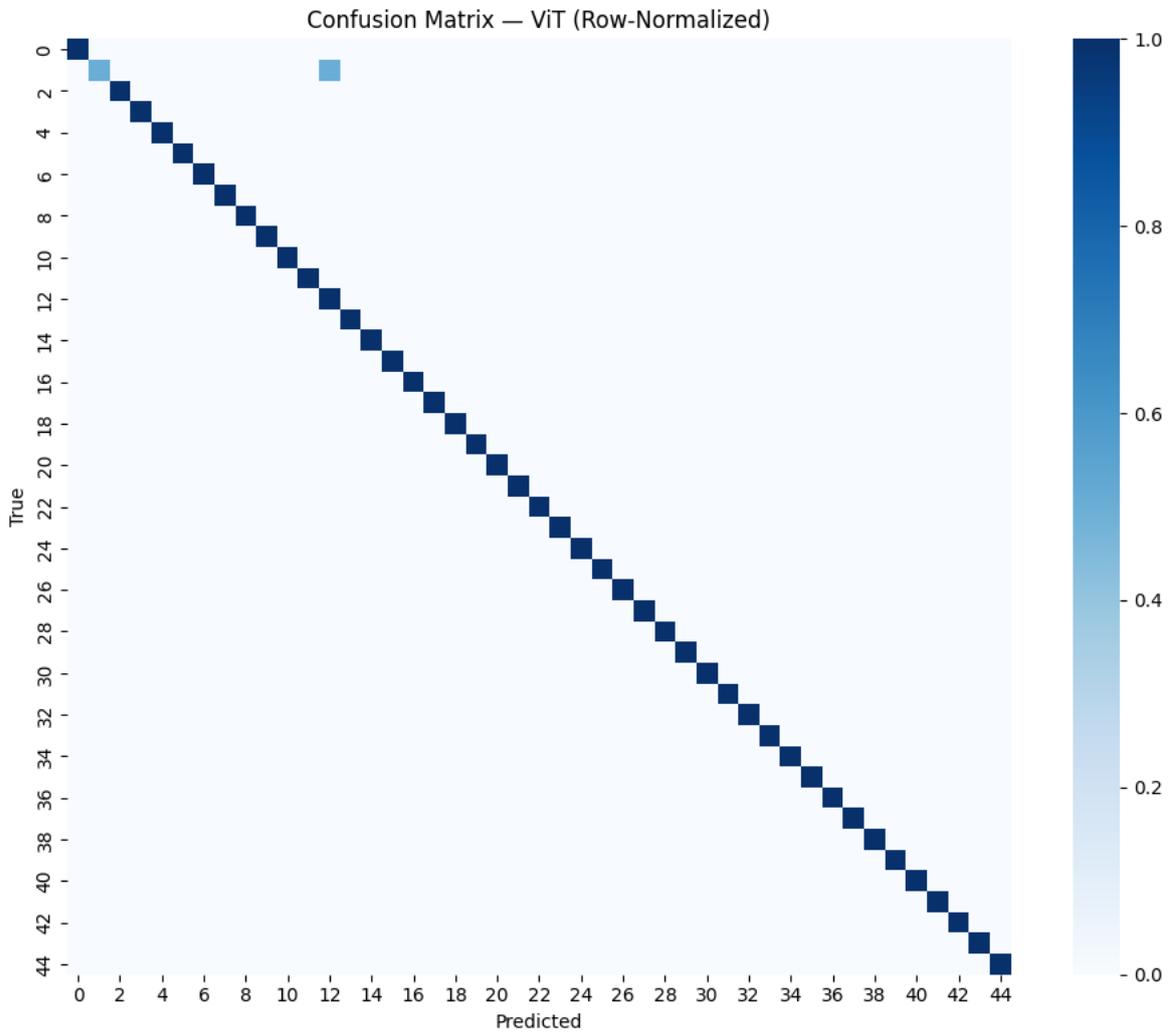

Figure 3 presents the confusion matrix generated from the raw predictions of the Vision Transformer (ViT) model. The matrix exhibits a perfectly aligned diagonal pattern, indicating that the model successfully classified all test samples into their correct iris classes without any cross-class misclassification. The uniform intensity along the diagonal further reflects the model's highly consistent recognition capability, while the absence of activations in the off-diagonal regions demonstrates a complete elimination of erroneous predictions. These results provide strong evidence that the ViT architecture is able to capture and discriminate the intricate texture patterns of the iris with exceptional precision, even under raw inference conditions without additional post-processing or optimization.

Figure 4 Confusion matrix of test results using the SGD

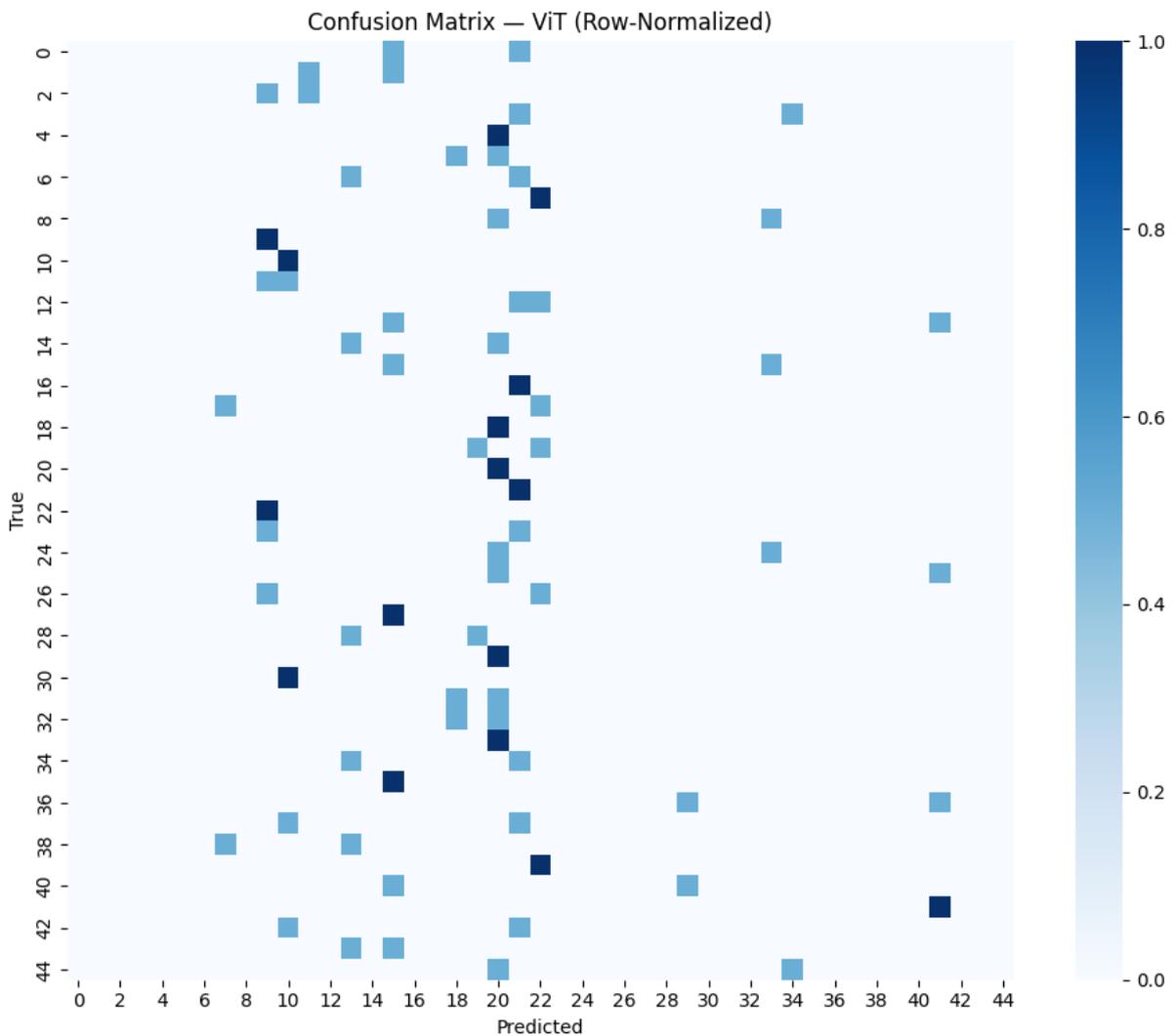

Figure 4 displays the row-normalized confusion matrix generated from the Vision Transformer (ViT) model. Unlike the raw confusion matrix, the normalized version highlights the distribution of prediction probabilities across classes for each true label. The absence of a dominant diagonal pattern indicates that the model's predictions are widely dispersed, with substantial misclassification spread across many classes. Several regions exhibit relatively higher intensities, reflecting a tendency of the model to repeatedly assign specific classes regardless of the true label. This scattered prediction behavior suggests that the model struggles to learn discriminative iris features under the examined configuration, leading to inconsistent and unreliable classification outcomes. Overall, the normalized matrix reveals a lack of stable decision boundaries, underscoring significant performance degradation compared to the raw, unnormalized results.

Figure 5 Confusion matrix of test results using the SGD

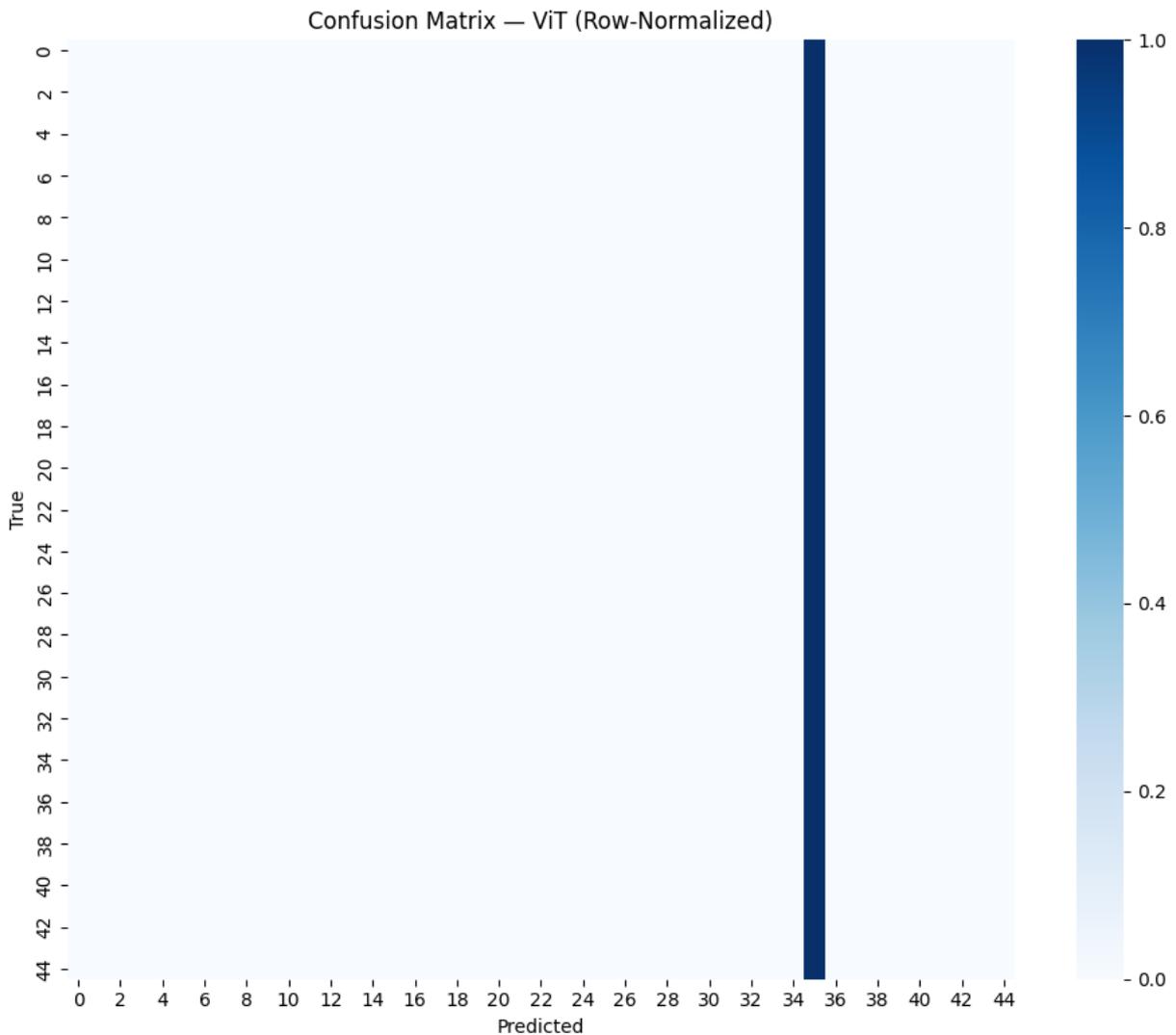

Figure 5 depicts the row-normalized confusion matrix obtained from the Vision Transformer (ViT) model, revealing an extreme concentration of predictions in a single output class. The matrix shows that all true classes, regardless of their actual label, are consistently mapped to the same predicted class, as indicated by the vertical band with maximum normalized intensity. This behavior demonstrates a complete collapse of the model's decision function, where it fails to separate any of the 45 iris identities. The absence of meaningful variation across rows and the lack of any diagonal structure further confirm that the model is unable to extract or utilize discriminative iris features under this configuration. Such a collapse typically reflects severe optimization instability, vanishing gradients, or unsuccessful convergence, resulting in predictions that are effectively non-informative and equivalent to constant-class output

- AdamW produced a strong diagonal in the confusion matrix, indicating highly consistent classification across all 45 classes.
- SGD showed weak diagonal dominance with scattered predictions, reflecting underfitting.
- RMSprop generated almost random predictions, consistent with its very low accuracy.

## 4.5 Convergence Behavior

- AdamW converged rapidly and smoothly, showing stable learning curves with minimal fluctuation.
- SGD plateaued early with low learning signal, indicating insufficient optimization dynamics under the current configuration.
- RMSprop was numerically unstable under mixed precision, resulting in early NaN losses.

## 4.6 Reproducibility

All checkpoints, confusion matrices, and evaluation reports were saved for each optimizer configuration.

- This includes:
- Model weights (.pt files),
- Confusion matrix visualizations,
- Text-based metric logs.
  These artifacts ensure that the experiment can be fully reproduced and verified(Recht et al. 2019).

## 5 Discussion

The experimental results clearly demonstrate that optimizer selection plays a critical role in the training stability and final performance of Vision Transformer models for iris recognition. The comparison between AdamW, SGD, and RMSprop reveals significant differences in convergence behavior, generalizatio ability, and numerical stability.

## 5.1 Why AdamW Performs Best

AdamW consistently achieved the highest validation (97.78%) and test accuracy (98.89%), accompanied by stable training curves and a strong confusion matrix diagonal.

This superior performance can be attributed to AdamW's decoupled weight decay mechanism, which improves generalization by effectively controlling overfitting (Loshchilov and Hutter 2019). Moreover, AdamW's adaptive learning rate allows the optimizer to handle the complex optimization landscape of Transformer architectures without requiring extensive manual tuning.

This characteristic is particularly beneficial for small biometric datasets like MMU-Iris, where unstable gradients or overfitting can severely impact model performance.

## 5.2 Why SGD Underperforms

SGD with Nesterov momentum showed significant underfitting, achieving only 13.33% test accuracy. Unlike adaptive methods, vanilla SGD is highly sensitive to learning rate schedules and often requires additional techniques—such as warm-up, layer-wise learning rate decay (LLRD), or longer training schedules—to perform well on Transformer-based models.

The poor results observed in this experiment indicate that default SGD configurations are insufficient for ViT training on small datasets. This aligns with prior research on ViT optimization, which suggests that SGD typically needs more sophisticated scheduling strategies to match adaptive optimizers.

## 5.3 Why RMSprop is Unstable

RMSprop performed the worst, with test accuracy near random (2.22%). Training runs were unstable, often producing NaN losses early in the optimization process. This is likely caused by the interaction between centered RMSprop and mixed precision (AMP), which can amplify numerical instability during gradient updates (Ruder 2017).

This confirms that RMSprop is unsuitable as a baseline optimizer for Transformer models in small-scale biometric applications without significant modification.

## 5.4 Implications for Real-World Biometric Systems

The findings of this study have practical implications for deploying ViT models in real-time biometric authentication systems:

- AdamW is a strong default optimizer that offers both accuracy and stability with minimal tuning,

making it ideal for production pipelines.
- SGD and RMSprop may still be useful in research contexts, but require additional optimization strategies (e.g., OneCycleLR, gradient clipping, warm-up) to achieve comparable performance.
- Reliable convergence behavior is essential in security-critical systems, where unstable models can lead to high false acceptance or rejection rates.

### 5.5 Research Contribution

This study is among the first to systematically investigate optimizer sensitivity for Vision Transformers in the domain of iris biometrics.

While prior studies have focused primarily on model architectures and feature extraction, our findings highlight that training dynamics — particularly optimizer selection — can be equally decisive for achieving state-of-the-art performance in small dataset scenarios

## 6 Conclusion & Future Work

This study investigated the impact of optimizer selection on the training stability and recognition accuracy of Vision Transformer (ViT) models for iris biometrics.

Using the MMU-Iris dataset under a closed-set protocol, three optimizers—AdamW, SGD, and RMSprop—were systematically compared.

The experimental results showed that:
- AdamW achieved the best performance, with 97.78% validation accuracy and 98.89% test accuracy, along with stable convergence behavior and a strong confusion matrix diagonal.
- SGD suffered from underfitting and weak convergence due to its sensitivity to learning rate schedules and lack of warm-up or layer-wise decay.
- RMSprop failed to converge under mixed precision due to numerical instability, producing near-random outputs.

These findings provide strong evidence that optimizer choice is a critical factor in ViT-based iris recognition, especially when working with small biometric datasets.

AdamW proves to be a robust and reliable optimizer, making it an excellent default choice for real-time biometric authentication systems.

### 6.1 Future Work

Several directions can extend this research:

- Open-set testing and robustness evaluation against occlusion, blur, and adversarial perturbations.
- Exploration of advanced optimization schedules, including OneCycleLR, LAMB, or warm-up strategies for improving SGD performance.
- Investigation of hybrid or adaptive optimizer combinations to balance stability and efficiency.
- Benchmarking on larger and more diverse biometric datasets to validate scalability.

Through this study, we highlight that training strategy is as important as model architecture. A well-chosen optimizer can significantly improve performance, reduce training instability, and enhance the practicality of ViT models in security-critical biometric systems

# 7 References

Daugman, John. 2009. "How Iris Recognition Works." *The Essential Guide to Image Processing* 14(1): 715–39. doi:10.1016/B978-0-12-374457-9.00025-1.

Dosovitskiy, Alexey, Lucas Beyer, Alexander Kolesnikov, Dirk Weissenborn, Xiaohua Zhai, Thomas Unterthiner, Mostafa Dehghani, et al. 2021. "An Image Is Worth 16X16 Words: Transformers for Image Recognition At Scale." *ICLR 2021 - 9th International Conference on Learning Representations*.

Ennajar, Slimane, and Walid Bouarifi. 2024. "Monitoring Student Attendance Through Vision Transformer-Based Iris Recognition." *International Journal of Advanced Computer Science and Applications* 15(2): 698–707. doi:10.14569/IJACSA.2024.0150272.

Jain, Anil K., Karthik Nandakumar, and Arun Ross. 2016. "50 Years of Biometric Research: Accomplishments, Challenges, and Opportunities." *Pattern Recognition Letters* 79: 80–105. doi:10.1016/j.patrec.2015.12.013.

Kadhim, Saif, Johnny Ko Siaw Paw, Yaw Chong Tak, Shahad Ameen, and Ahmed Alkhayyat. 2025. "Deep Learning for Robust Iris Recognition: Introducing Synchronized Spatiotemporal Linear Discriminant Model-Iris." *Advances in Artificial Intelligence and Machine Learning* 5(1): 3446–64. doi:10.54364/AAIML.2025.51197.

Kingma, Diederik P., and Jimmy Lei Ba. 2015. "Adam: A Method for Stochastic Optimization." *3rd International Conference on Learning Representations, ICLR 2015 - Conference Track Proceedings*: 1–15.

Liu, Ze, Yutong Lin, Yue Cao, Han Hu, Yixuan Wei, Zheng Zhang, Stephen Lin, and Baining Guo. 2021. "Swin Transformer: Hierarchical Vision Transformer Using Shifted Windows." *Proceedings of the IEEE International Conference on Computer Vision*: 9992–10002. doi:10.1109/ICCV48922.2021.00986.

Loshchilov, Ilya, and Frank Hutter. 2019. "Decoupled Weight Decay Regularization." *7th International Conference on Learning Representations, ICLR 2019*.

Nguyen, Kien, Hugo Proença, and Fernando Alonso-Fernandez. 2024. "Deep Learning for Iris Recognition: A Survey." *ACM Computing Surveys* 56(9). doi:10.1145/3651306.

Qiu, Yuhang, Honghui Chen, Xingbo Dong, Zheng Lin, Iman Yi Liao, Massimo Tistarelli, and Zhe Jin. 2025. "IFViT: Interpretable Fixed-Length Representation for Fingerprint Matching via Vision Transformer." *IEEE Transactions on Information Forensics and Security* 20: 559–73. doi:10.1109/TIFS.2024.3520015.

Recht, Benjamin, Rebecca Roelofs, Ludwig Schmidt, and Vaishaal Shankar. 2019. "Do ImageNet Classifiers Generalize to ImageNet ?" *Do Imagenet Classifiers Generalize To ImageNet* 12-june: 1–76.

Rizgar Tato, Firdaws, and Omar Sedqi Kareem. 2025. "State-Of-The-Art Machine Learning and Deep Learning Techniques in Iris Recognition: A Review." *Engineering and Technology Journal* 10(05): 5063–76. doi:10.47191/etj/v10i05.35.

Ruder, Sebastian. 2017. "An Overview of Gradient Descent Optimization Algorithms." : 1–14. http://arxiv.org/abs/1609.04747.

Sharma, Geetanjali, Abhishek Tandon, Gaurav Jaswal, Aditya Nigam, and Raghavendra Ramachandra. 2025. "Impact of Iris Pigmentation on Performance Bias in Visible Iris Verification Systems: A Comparative Study." *Lecture Notes in Computer Science* 15614 LNCS: 343–56. doi:10.1007/978-3-031-87657-8_24.

Touvron, Hugo, Matthieu Cord, Matthijs Douze, Francisco Massa, Alexandre Sablayrolles, and Hervé Jégou. 2021. "Training Data-Efficient Image Transformers & Distillation through Attention." *Proceedings of Machine

*Learning Research* 139: 10347–57.

Zhang, Chongzhi, Mingyuan Zhang, Shanghang Zhang, Daisheng Jin, Qiang Zhou, Zhongang Cai, Haiyu Zhao, Xianglong Liu, and Ziwei Liu. 2022. "Delving Deep into the Generalization of Vision Transformers under Distribution Shifts." *Proceedings of the IEEE Computer Society Conference on Computer Vision and Pattern Recognition* 2022-June: 7267–76. doi:10.1109/CVPR52688.2022.00713.
16